\documentclass[letterpaper, 10 pt, journal, twoside]{IEEEtran}
\usepackage{cite}
\usepackage{amsmath,amssymb,amsfonts}
\usepackage{algorithmic}
\usepackage{graphicx}
\usepackage{textcomp}
\usepackage{xcolor}
\usepackage{multirow}
\usepackage{makecell}
\ifCLASSINFOpdf
\else
\fi
\begin{document}
%
\title{SR-LIVO: LiDAR-Inertial-Visual Odometry and Mapping with Sweep Reconstruction}
%
%
%

\author{Zikang~Yuan$^{1}$, Jie~Deng$^{2}$, Ruiye~Ming$^{2}$, Fengtian~Lang$^{2}$ and Xin~Yang$^{2*}$
	\thanks{$^{1}$Zikang~Yuan is with Institute of Artificial Intelligence, Huazhong University of Science and Technology, Wuhan, 430074, China. (E-mail: {\tt\small yzk2020@hust.edu.cn})}%
	\thanks{$^{2}$Jie~Deng, Ruiye~Ming, Fengtian~Lang and Xin~Yang$^{*}$ are with the Electronic Information and Communications, Huazhong University of Science and Technology, Wuhan, 430074, China. (* represents the corresponding author. E-mail: {\tt\small u202013949@hust.edu.cn; M202272555@hust.edu.cn; M202372913@hust.edu.cn; xinyang2014@hust.edu.cn})}%
}
%
%

\markboth{IEEE Robotics and Automation Letters. Preprint Version. Accepted Month, Year}
{FirstAuthorSurname \MakeLowercase{\textit{et al.}}: ShortTitle} 

%



\maketitle

\begin{abstract}
Existing LiDAR-inertial-visual odometry and mapping (LIV-SLAM) systems mainly utilize the LiDAR-inertial odometry (LIO) module for structure reconstruction and the visual-inertial odometry (VIO) module for color rendering. However, the accuracy of VIO is often compromised by photometric changes, weak textures and motion blur, unlike the more robust LIO. This paper introduces SR-LIVO, an advanced and novel LIV-SLAM system employing sweep reconstruction to align reconstructed sweeps with image timestamps. This allows the LIO module to accurately determine states at all imaging moments, enhancing pose accuracy and processing efficiency. Experimental results on two public datasets demonstrate that: 1) our SR-LIVO outperforms existing state-of-the-art LIV-SLAM systems in both pose accuracy and time efficiency; 2) our LIO-based pose estimation prove more accurate than VIO-based ones in several mainstream LIV-SLAM systems (including ours). We have released our source code to contribute to the community development in this field.
\end{abstract}


%
\IEEEpeerreviewmaketitle

\section{Introduction}
\label{Introduction}

\IEEEPARstart{I}{n} robotic applications like autonomous vehicles \cite{levinson2011towards} and drones \cite{gao2019flying, kong2021avoiding}, cameras, 3D light detection and ranging (LiDAR) and inertial measurement unit (IMU) are key sensors. The integration of IMU measurements can provide motion prior to ensure accurate and quick state estimation. While LiDAR excels in capturing 3D structures, it lacks color information which cameras compensate for. This synergy has led to the rise of LiDAR-inertial-visual odometry and mapping (LIV-SLAM) as a leading method for accurate state estimation and dense color map reconstruction.

\begin{figure}
	\begin{center}
		\includegraphics[scale=0.5]{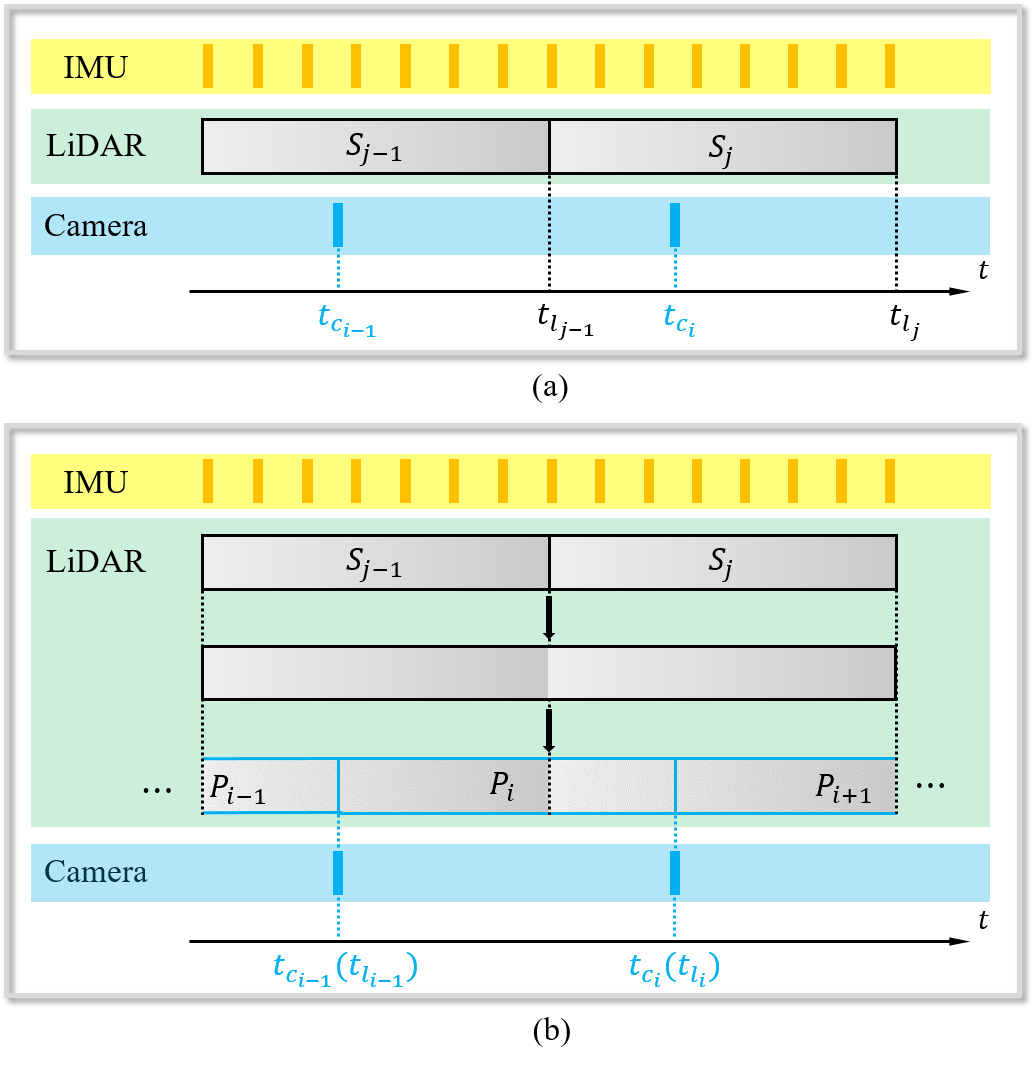}
		\caption{Illustration of (a) the raw sensor measurements from IMU, LiDAR and camera, (b) the raw sensor measurements from IMU, camera and the reconstructed LiDAR sweep data, where the end timestamp of reconstructed sweep is aligned with the timestamp of captured image.}
		\label{fig1}
	\end{center}
\end{figure}

Existing state-of-the-art LIV-SLAM systems \cite{lin2021r, lin2022r, zheng2022fast} mainly consist of a LIO module and a VIO module. The LIO module reconstructs 3D structures while the VIO module colors them. Both LIO and VIO modules perform state estimation. The LIO module obtains the estimated state at the end timestamp of a LiDAR sweep (e.g., $t_{l_{j-1}}$ and $t_{l_j}$ in Fig. \ref{fig1} (a)), and the VIO module solves the state at the timestamp of each captured image (e.g., $t_{c_{i-1}}$ and $t_{c_i}$ in Fig. \ref{fig1} (a)). Compared to LIO, VIO is less reliable to photometric changes, weak textures and motion blur, and thus suffers less accurate pose results and in turn degrade colored reconstruction.

In this paper, we propose SR-LIVO, a novel and advanced LVI-SLAM framework that enhances both accuracy and reliability. We shift state estimation entirely to the LIO module, known for its better precision. A key challenge brought by this design is that: the timestamp of the captured images and the end timestamp of LiDAR sweeps are often not aligned. Without the state of the image acquisition moment, we can hardly use image data to render color for the restored 3D structures. To address this issue, we adopt the sweep reconstruction method proposed in our previous work \cite{yuan2023sdv} for data synchronization (e.g., $t_{l_{i-1}}$ and $t_{l_i}$ in Fig. \ref{fig1} (b)) with the timestamp of captured images (e.g., $t_{c_{i-1}}$ and $t_{c_i}$ in Fig. \ref{fig1} (b)). In this way, the LIO module can directly estimate states at image capture moments. Consequently, the vision module's role is simplified to optimizing camera parameters (e.g., camera intrinsics, extrinsics and time-offset) and completing the color rendering task.

Experimental results on the public datasets $NTU\_VIRAL$ \cite{nguyen2022ntu}, $R3Live$ \cite{lin2022r} demonstrate the following key findings: 1) Our system outperforms existing state-of-the-art LIO systems (i.e., \cite{lin2021r, lin2022r}) in terms of the smaller absolute trajectory error (ATE), and is much more efficient than \cite{lin2022r}; 2) The estimated pose from the LIO module is more accurate than that from the VIO modules of several mainstream LIV-SLAM systems. Meanwhile, the visualization results demonstrate that our SR-LIVO can achieve comparable reconstruction results to \cite{lin2022r} on their self-collected dataset (i.e., $R3Live$), and can achieve much superior results to \cite{lin2022r} on the $NTU\_VIRAL$ dataset.

To summarize, the main contributions of this work are three folds: 1) We identify an important function of sweep reconstruction, i.e., aligning LiDAR sweep and image timestamps; 2) We design a new LIV-SLAM system based on sweep reconstruction, where the vision module is no longer needed to perform state estimation but only for coloring the reconstructed map; 3) We have released the source code of our system to benefit the development of the community\footnote{https://github.com/ZikangYuan/sr\_livo}.

The rest of this paper is structured as follows. Sec. \ref{Related Work} reviews the relevant literature. Sec. \ref{Preliminary} provides preliminaries. Secs. \ref{System Overview} and \ref{System Details} presents system overview and details, followed by experimental evaluation in Sec. \ref{Experiments}. Sec. \ref{Experiments} concludes the paper.

\section{Related Work}
\label{Related Work}

In recent years, various LiDAR-visual and LiDAR-inertial-visual fusion frameworks have been proposed. V-LOAM \cite{zhang2018laser} is the first cascaded LIV-SLAM framework that provides motion priori for LiDAR odometry via a loosely coupled VIO. \cite{shao2019stereo} cascades a tightly-coupled stereo VIO, a LiDAR odometry and a LiDAR-based loop closing module together. Compared with \cite{zhang2018laser} and \cite{shao2019stereo}, the vision module of DV-LOAM \cite{wang2021dv} utilizes the direct method to perform pose estimation and multi-frame joint optimization in turn to make the vision module provide more accurate motion priori for the subsequent LiDAR module. Lic-Fusion \cite{zuo2019lic} combines the IMU measurements, sparse visual features, LiDAR features with online spatial and temporal calibration within the multi state constrained Kalman filter (MSCKF) framework. To further enhance the accuracy and the robustness of the LiDAR points registration, LIC-Fusion 2.0 \cite{zuo2020lic} proposes a plane-feature tracking algorithm across multiple LiDAR sweeps in a sliding-window and refines the pose of sweep within the window. LVI-SAM \cite{shan2021lvi} integrates the data from camera, LiDAR and IMU into a tightly-coupled graph-based optimization framework. The vision and LiDAR module of LVI-SAM can run independently when each other fails, or jointly when both visual and LiDAR features are sufficient. R2Live \cite{lin2021r} firstly proposes to run a LIO module and a VIO module in parallel, where the LIO module provides geometric structure information for the VIO module. The back end utilizes the visual landmarks to perform graph-based optimization. Based on R2Live, R3Live \cite{lin2022r} omits the graph-based optimization module and adds a color rendering module for dense color map reconstruction. Compared to R3Live, Fast-LIVO \cite{zheng2022fast} combines LiDAR, camera and IMU measurements into a single error state iterated Kalman filter (ESIKF), which can be updated by both LiDAR and visual observations. mVIL-Fusion \cite{wang2022mvil} proposes a three-staged cascading LVI-SLAM framework, which consists of a LiDAR-assisted VIO module, a multi sweep-to-sweep joint optimization module, a sweep-to-map optimization module and a loop closing module. VILO-SLAM \cite{peng2023vilo} fuses 2D LiDAR residual factor, IMU residual factor and visual reprojection residual factor into an optimization-based framework. In \cite{yin2023solid}, a weight function is designed based on geometric structure and reflectivity to improve the performance of solid-state LIO under severe linear acceleration and angular velocity changes.

\section{Preliminary}
\label{Preliminary}

\subsection{Coordinate Systems}
\label{Coordinate Systems}

We denote $(\cdot)^w$, $(\cdot)^l$, $(\cdot)^o$ and $(\cdot)^c$ as a 3D point in the world coordinate, the LiDAR coordinate, the IMU coordinate and the camera coordinate respectively. The world coordinate is coinciding with $(\cdot)^o$ at the starting position.

We denote the IMU coordinate for taking the $i_{th}$ IMU measurement at time $t_i$ as $o_i$ and the corresponding camera coordinate at $t_i$ as $c_i$. The transformation matrix (i.e., extrinsics) from $c_i$ to $o_i$ is denoted as $\mathbf{T}_{c_i}^{o_i} \in S E(3)$, where $\mathbf{T}_{c_i}^{o_i}$ consists of a rotation matrix $\mathbf{R}_{c_i}^{o_i} \in S O(3)$ and a translation vector $\mathbf{t}_{c_i}^{o_i} \in \mathbb{R}^3$. Similarly, we can also obtain the extrinsics form LiDAR to IMU, i.e., $\mathbf{T}_{l_i}^{o_i}$. For the datasets involved in this work, we use the internal IMU of LiDAR, therefore, we treat $\mathbf{T}_{l_i}^{o_i}$ as absolutely accurate and do not require online correction. For $\mathbf{T}_{c_i}^{o_i}$, we optimize it online because the camera is a separate external sensor.

\subsection{Distortion Correction}
\label{Distortion Correction}

For each camera image, we utilized the offline-calibrated distortion parameters to correct the image distortion. For LiDAR points, we utilize the IMU-integrated pose to compensate the motion distortion.

\section{System Overview}
\label{System Overview}

\begin{figure}
	\begin{center}
		\includegraphics[scale=0.5]{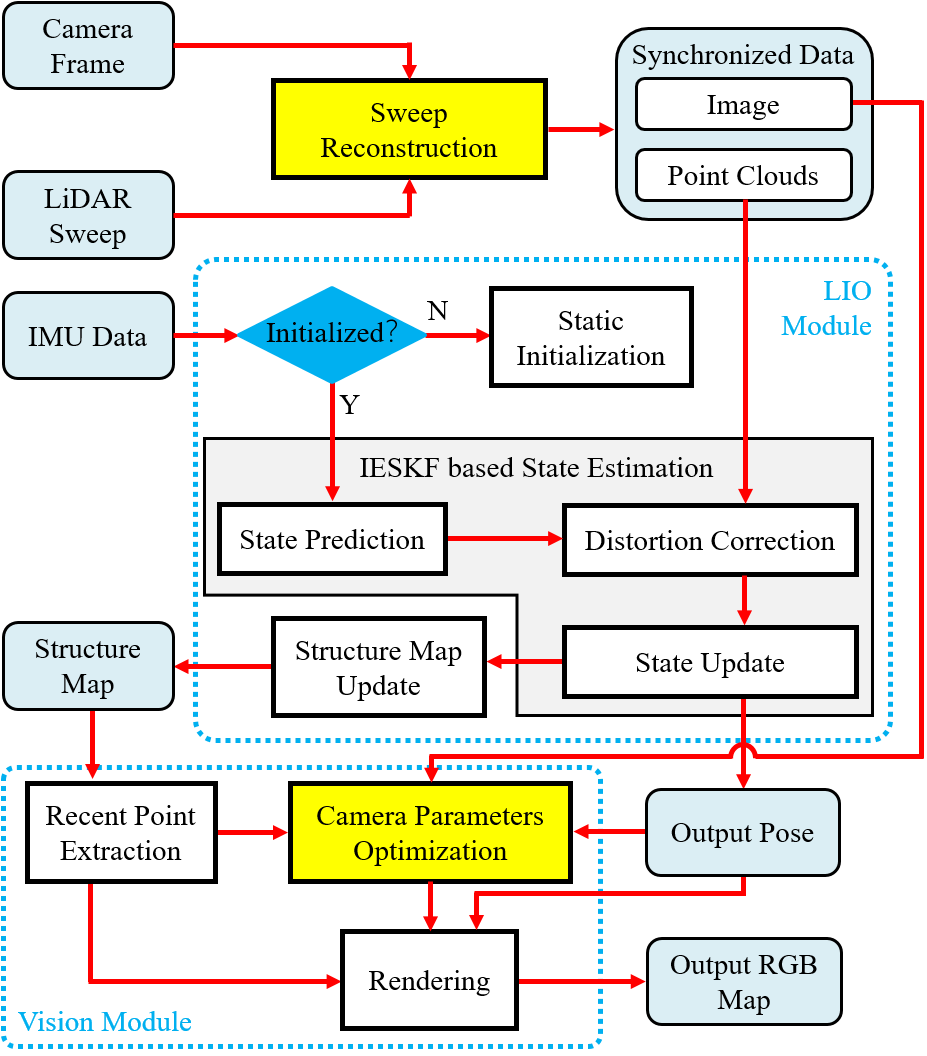}
		\caption{Overview of our SR-LIVO which consists of three main modules: a sweep reconstruction module for timestamp alignment, a LIO module for state estimation and structure restoring, and a vision module for camera parameters optimization and color rendering. The yellow rectangles indicate the operations of our system that are different from that of existing state-of-the-art frameworks.}
		\label{fig2}
	\end{center}
\end{figure}

Fig. \ref{fig2} illustrates the framework of our system which consists of three main modules: a sweep reconstruction module, a LIO state estimation module and a vision module. The sweep reconstruction module aligns the end timestamp of reconstructed sweep with the timestamp of captured image. The LIO module estimates the state of the hardware platform, and restores the structure in real time. The vision module optimizes the camera parameters including camera intrinsics, extrinsics, time-offset, and renders color to the restored structure map in real time. For map management, we utilized the Hash voxel map, which is the same as CT-ICP \cite{dellenbach2022ct}. The implementation details of various parts of the LIO module are exactly the same as our previous work SR-LIO \cite{yuan2022sr}, therefore, we omit the introduction of this module, and only introduce the details of sweep reconstruction and our vision module in Sec. \ref{System Details}.

\section{System Details}
\label{System Details}

\subsection{Sweep Reconstruction}
\label{Sweep Reconstruction}

In our previous work SDV-LOAM \cite{yuan2023sdv}, we firstly propose the sweep reconstruction idea, which increases the frequency of LiDAR sweeps to the same frequency as camera images. In this work, we point out another important function of this idea: aligning the end timestamp of reconstructed sweep to the timestamp of captured image. According to the frequency of LiDAR sweeps and camera images, the processing method is also different in practice. There are three specific cases as following:

\begin{figure}
	\begin{center}
		\includegraphics[scale=0.5]{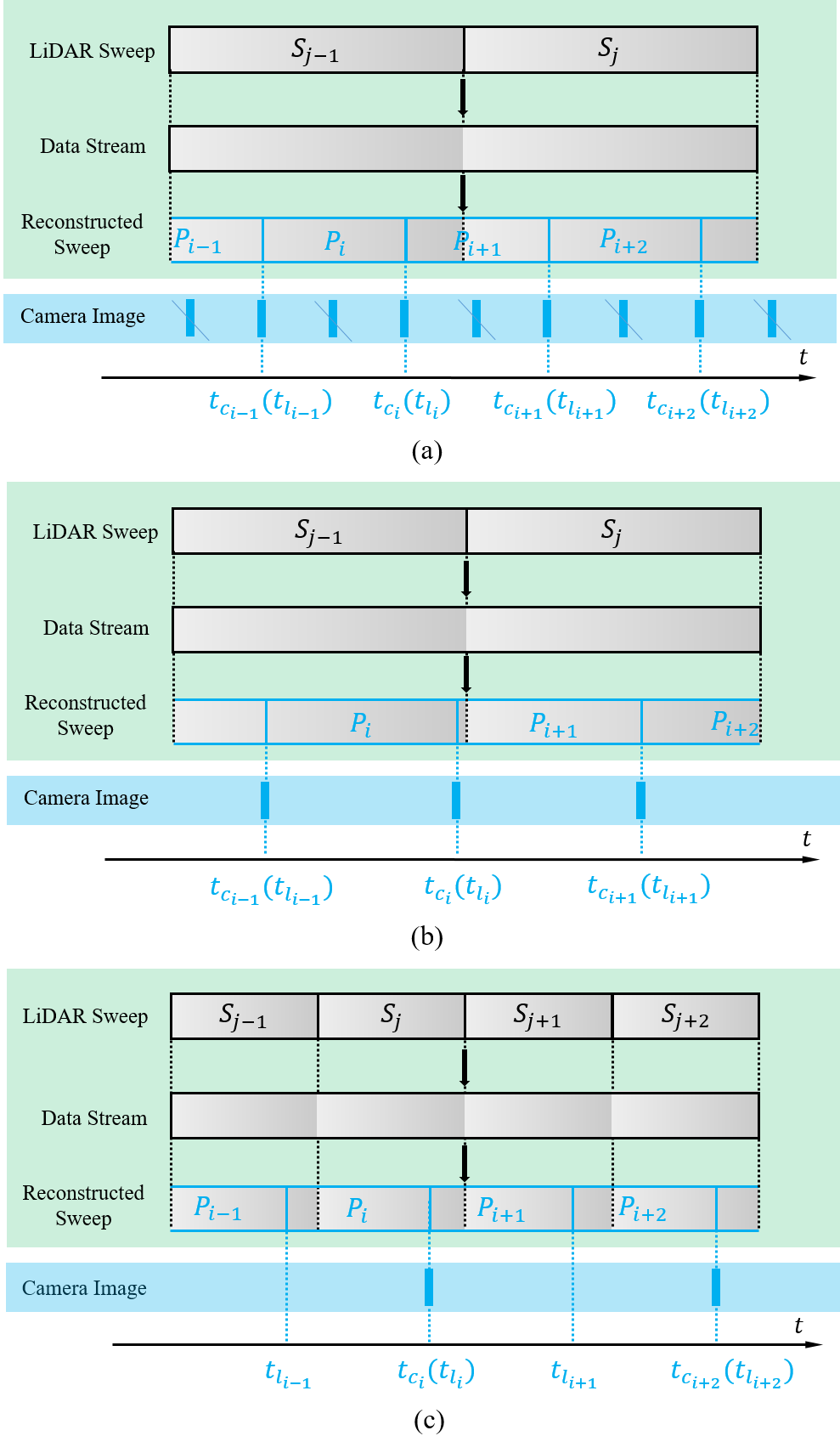}
		\caption{Illustration of the sweep reconstruction method under three situations: (a) The frequency of captured images is more than twice that of raw LiDAR sweeps. (b) The frequency of captured images is less than twice but more than that of raw LiDAR sweeps. (c) The frequency of captured images is less than that of raw LiDAR sweeps.}
		\label{fig3}
	\end{center}
\end{figure}

\textbf{The frequency of captured images is more than twice that of raw LiDAR sweeps (as shown in Fig. \ref{fig3} (a)).} In this case, we firstly down-sample the frequency of camera images to twice the frequency of raw LiDAR sweeps. Then we disassemble the raw LiDAR sweep (i.e., $S_{j-1}$ and $S_j$ in Fig. \ref{fig3} (a)) into continuous point cloud data stream, and recombine the point cloud data stream with aligning the end timestamp (e.g., $t_{l_{i}}$ in Fig. \ref{fig3} (a)) of reconstructed sweep (i.e., $P_i$ in Fig. \ref{fig3} (a)) to the timestamp of down-sampled captured image (e.g., $t_{c_i}$ in Fig. \ref{fig3} (a)). In this way, the number of LiDAR points in reconstructed sweep is only half that in raw input sweep. For spinning LiDAR, not only the number, but also the horizontal distribution range is reduced from $360^{\circ}$ to $180^{\circ}$. If the frequency of reconstructed sweeps is more than twice that of raw sweeps, the number and horizontal distribution range of points will decrease further, and eventually the LIO module may fail to perform properly. Therefore, we need to down-sample the frequency of camera images to at most twice the frequency of raw LiDAR sweeps.

\textbf{The frequency of captured images is less than twice but more than that of raw LiDAR sweeps (as shown in Fig. \ref{fig3} (b)).} In this case, we directly disassemble the raw LiDAR sweep (i.e., $S_{j-1}$ and $S_j$ in Fig. \ref{fig3} (b)) into continuous point cloud data stream, and recombine the point cloud data stream with aligning the end timestamp (e.g., $t_{l_{i}}$ in Fig. \ref{fig3} (b)) of reconstructed sweep (e.g., $P_i$ in Fig. \ref{fig3} (b)) to the timestamp of captured image (e.g., $t_{c_i}$ in Fig. \ref{fig3} (b)). In this way, the point cloud number and the horizontal distribution range of reconstructed sweeps are less than that of raw LiDAR sweeps, but also enough to support the LIO module run properly.

\textbf{The frequency of captured images is less than that of raw LiDAR sweeps (as shown in Fig. \ref{fig3} (c)).} In this case, we directly disassemble the raw LiDAR sweep (i.e., $S_{j-1}$, $S_j$, $S_{j+1}$ and $S_{j+2}$ in Fig. \ref{fig3} (c)) into continuous point cloud data stream, and recombine the point cloud data stream according to the following two situations: 1) We assume the current reconstructed sweep begins at $t_{l_i}$ but the end timestamp is not yet determined. When the time interval from current moment to $t_{l_i}$ reaches the time period of a raw LiDAR sweep and there are no images around current moment, we set the current moment as the end timestamp (e.g., $t_{l_{i+1}}$ in Fig. \ref{fig3} (c)) of this reconstructed sweep (e.g., $P_{i+1}$ in Fig. \ref{fig3} (c)). 2) We assume the current reconstructed sweep begins at $t_{l_{i+1}}$ but the end timestamp is not yet determined. When the current moment reaches the timestamp of captured image (e.g., $t_{c_{i+2}}$ in Fig. \ref{fig3} (c)) and there is a sufficient time interval from $t_{c_{i+2}}$ to $t_{l_{i+1}}$, we set the current moment as the end timestamp (e.g., $t_{c_{i+2}}$($t_{l_{i+2}}$) in Fig. \ref{fig3} (c)) of reconstructed sweep (e.g., $P_{i+2}$ in Fig. \ref{fig3} (c)). Under the situation 2), not all synchronized data includes image data (i.e., $P_{i-1}$ and $P_{i+1}$ in Fig. \ref{fig3} (c)). For synchronized data without camera images, we just utilize the LIO module to estimate state without the vision module running. For synchronized data with both point cloud and image data, the LIO module and the vision module execute in turn.

\subsection{Vision Module}
\label{Vision Module}

\subsubsection{Recent Point Extraction}
\label{Recent Point Extraction}

Similar as R3Live \cite{lin2022r}, firstly we record all recently visited voxels $\left\{V_1, V_2, \cdots, V_m\right\}$ when performing structure map update. Then, we select the newest added point from each visited voxel $V_i$, to obtain the recent point set $P_r = \left\{\mathbf{p}_1, \mathbf{p}_2, \cdots, \mathbf{p}_m\right\}$.

\subsubsection{Camera Parameters Optimization}
\label{Camera Parameters Optimization}

Different from existing state-of-the-art frameworks (e.g., R3Live and Fast-LIVO), we have estimated the state at the timestamp (e.g., $t_k$) of captured images (e.g., $c_k$) in the LIO module. Therefore, we no longer need to solve state (i.e., pose, velocity and IMU bias) in the vision module, but only need to utilize the image data to optimize some camera parameters:
\begin{equation}
\label{equation1}
	\boldsymbol{x}_k=\left[t_{c_k}^{o_k}, \mathbf{R}_{c_k}^{o_k}, \mathbf{t}_{c_k}^{o_k}, \phi_k\right]^T
\end{equation}
where $t_{c_k}^{o_k}$ is the time-offset between IMU and camera while LiDAR is assumed to be synced with the IMU already. $\mathbf{R}_{c_k}^{o_k}$ and $\mathbf{t}_{c_k}^{o_k}$ are the extrinsics between camera and IMU. $\phi_k=\left[f_{x_k}, f_{y_k}, c_{x_k}, c_{y_k}\right]^T$ are the camera intrinsics, where $(f_{x_k}, f_{y_k})$ denote the camera focal length, $(c_{x_k}, c_{y_k})$ denote the offsets of the principal point from the top-left corner of the image plane.

To balance the effects of previous estimates and the current image observation on camera parameters, we utilize an ESIKF to optimize the camera parameters $\boldsymbol{x}_k$, where the error state $\delta \boldsymbol{x}_k$ is defined as:
\begin{equation}
\label{equation2}
	\delta \boldsymbol{x}_k=\left[\delta t_{c_k}^{o_k}, \delta \mathbf{R}_{c_k}^{o_k}, \delta \mathbf{t}_{c_k}^{o_k}, \delta \phi_k\right]^T
\end{equation}
For the state prediction, the error state $\delta \boldsymbol{x}_k$ and covariance $\mathbf{P}_k$ is propagated as:
\begin{equation}
\label{equation3}
	\begin{gathered}
		\delta \boldsymbol{x}_k=\mathbf{0} \\
		\mathbf{P}_k=\mathbf{P}_{k-1}
	\end{gathered}
\end{equation}
For the state update, the minimizing PnP projection error and the minimizing photometric error are used to update $\delta \boldsymbol{x}$ in turn.

\textbf{The minimizing PnP projection error.} Assuming that we have tracked $n$ map points $P_t = \left\{\mathbf{p}_1, \mathbf{p}_2, \cdots, \mathbf{p}_n\right\}$, and their projection on image $c_{k-1}$ is $\left\{\boldsymbol{\rho}_{1_{k-1}}, \boldsymbol{\rho}_{2_{k-1}}, \cdots, \boldsymbol{\rho}_{n_{k-1}}\right\}$, we leverage the Lucas-Kanade optical flow to find out their locations in the current image $c_k$: $\left\{\boldsymbol{\rho}_{1_k}, \boldsymbol{\rho}_{2_k}, \cdots, \boldsymbol{\rho}_{n_k}\right\}$. For the exemplar point $\mathbf{p}_i \in P_t$, we calculate the re-projection error by:
\begin{equation}
\label{equation4}
	\begin{gathered}
		\mathbf{p}_{i}^{c_k}=\left(\mathbf{R}_{o_k}^w \cdot \mathbf{R}_{c_k}^{o_k}\right)^T \cdot \mathbf{p}_i^w-\mathbf{R}_{c_k}^{o_k T} \cdot \mathbf{t}_{c_k}^{o_k}-\left(\mathbf{R}_{o_k}^w \cdot \mathbf{R}_{c_k}^{o_k}\right)^T \cdot \mathbf{t}_{o_k}^w \\
		\mathbf{r}^{\mathbf{p}_i}=\boldsymbol{\rho}_{i_k}-\pi\left(\mathbf{p}_{i}^{c_k,}, \boldsymbol{x}_k\right)
	\end{gathered}
\end{equation}
where $\pi\left(\mathbf{p}_{i}^{c_k,}, \boldsymbol{x}_k\right)$ is computed as below:
\begin{equation}
\label{equation5}
	\begin{aligned}
		& \pi\left(\mathbf{p}_{i}^{c_k}, \boldsymbol{x}_k\right)=\left[f_{x_k} \cdot \frac{\left[\mathbf{p}_{i}^{c_k}\right]_x}{\left[\mathbf{p}_{i}^{c_k}\right]_z}+c_{x_k}, f_{y_k} \cdot \frac{\left[\mathbf{p}_{i}^{c_k}\right]_y}{\left[\mathbf{p}_{i}^{c_k}\right]_z}+c_{y_k}\right]^T \\
		& +\frac{t_{c_k}^{o_k}}{\Delta t_{k-1, k}}\left(\boldsymbol{\rho}_{i_k}-\boldsymbol{\rho}_{i_{k-1}}\right) \\
		&
	\end{aligned}
\end{equation}
where $\Delta t_{k-1, k}$ is the time interval between the last image $c_{k-1}$ and the current image $c_k$. In Eq. \ref{equation5}, the first item is the pin-hole projection function and the second one is the online-temporal correction factor \cite{qinonline}. We can express the observation matrix $\mathbf{h}$ as:
\begin{equation}
\label{equation6}
	\mathbf{h}=\left[\mathbf{r}^{\mathbf{p}_1 T}, \mathbf{r}^{\mathbf{p}_2 T}, \cdots, \mathbf{r}^{\mathbf{p}_n^T}\right]^T
\end{equation}
The corresponding Jacobian matrix of observation constraint $\mathbf{H}$ is calculated as:
\begin{equation}
\label{equation7}
	\begin{aligned}
		& \mathbf{H}=\left[\mathbf{H}_1^T, \mathbf{H}_2^T, \cdots, \mathbf{H}_n^T\right]^T \\
		& \mathbf{H}_i=\left[\begin{array}{llll}
			{\frac{\partial \mathbf{r}^{\mathbf{p}_i}}{\partial t_{c_k}^{o_k}}}^T & {\frac{\partial \mathbf{r}^{\mathbf{p}_i}}{\partial \mathbf{R}_{c_k}^{o_k}}}^T & {\frac{\partial \mathbf{r}^{\mathbf{p}_i}}{\partial \mathbf{t}_{c_k}^{o_k}}}^T & {\frac{\partial \mathbf{r}^{\mathbf{p}_i}}{\partial \phi_k}}^T
		\end{array}\right]^T \\
		&
	\end{aligned}
\end{equation}
\textbf{The minimizing photometric error.} For the exemplar point $\mathbf{p}_i \in P_r$, if $\mathbf{p}_i$ has been rendered the color intensity $\boldsymbol{\gamma}_i$, we firstly project $\mathbf{p}_i^{w}$ to $(\cdot)^{c_k}$ by Eq. \ref{equation4} and \ref{equation5}, and then calculate the photometric error by:
\begin{equation}
\label{equation8}
	\mathbf{r}^{\mathbf{p}_i}=\boldsymbol{\gamma}_i-\boldsymbol{I}\left(\pi\left(\mathbf{p}_i^{c_k}, \boldsymbol{x}_k\right)\right)
\end{equation}
where $\boldsymbol{I}(\cdot)$ represents the color intensity of image pixel. The observation matrix and the corresponding Jacobin matrix have the similar formula as Eq. \ref{equation6}-\ref{equation7}.

The same as R3Live \cite{lin2022r}, for each image $c_k$, we firstly utilize \textbf{the minimizing PnP projection error} to update the ESIKF, and then utilize \textbf{the minimizing photometric error} to update the ESIKF. The difference is that the ESIKF of R3Live's vision module estimates both state and camera parameters of $c_k$, but we only optimize camera parameters of $c_k$, while the state of $c_k$ has been solved in our LIO module.

\subsubsection{Rendering}
\label{Rendering}

After the camera parameters have been optimized, we perform the rendering function to update the color of map points. To ensure the density of colored point cloud map, we not only render points in the recent point set $P_r$, but also render all points in recently visited voxels $\left\{V_1, V_2, \cdots, V_m\right\}$. The rendering function we utilize is the same as R3Live \cite{lin2022r}.

\section{Experiments}
\label{Experiments}

We evaluate our system on the drone-collected dataset $NTU\_VIRAL$ \cite{nguyen2022ntu} and the handheld device-collected dataset $R3Live$ \cite{lin2022r}. The sensors used in $NTU\_VIRAL$ dataset are the left camera, the horizontal 16-channel OS1 gen1$^{4}$ LiDAR and its internal IMU, while the high-accuracy laser tracking methods are employed to provide position ground truth. The sensors used in $R3Live$ dataset are the camera, the LiVOX AVAI LiDAR and its internal IMU, while no position ground truth data are provided. We utilize all sequences of $NTU\_VIRAL$ \cite{nguyen2022ntu} and 6 sequences (i.e., $r3live\_01$: $hku\_campus\_seq\_00$, $r3live\_02$: $hku\_campus\_seq\_01$, $r3live\_03$: $hku\_campus\_seq\_02$, $r3live\_04$: $hku\_park\_00$, $r3live\_05$: $hku\_park\_01$ and $r3live\_06$: $hkust\_campus\_seq\_02$) of $R3Live$ for evaluation. A consumer-level computer equipped with an Intel Core i7-11700 and 32 GB RAM is used for all experiments.

\subsection{Comparison of the State-of-the-Arts}
\label{Comparison of the State-of-the-Arts}

We compare our system with two state-of-the-art LIV-SLAM systems, i.e., R3Live \cite{lin2022r} and Fast-LIVO \cite{zheng2022fast}, on $NTU\_VIRAL$ dataset \cite{nguyen2022ntu} and $R3Live$ dataset \cite{lin2022r}. For the $NTU\_VIRAL$ dataset, we utilize the universal evaluation metrics - absolute translational error (ATE) as the evaluation metrics. The $R3Live$ dataset does not provide the position groundtruth, however, they return to their starting point at the end of most sequences. Therefore, we utilize the end-to-end error instead. For a fair comparison, we obtain the results of above systems based on the source code provided by the authors.

\begin{table}[]
	\begin{center}
		\caption{RMSE of ATE on NTU-VIRAL dataset (unit: m)}
		\label{table1}
			\begin{tabular}{p{1.6cm}<{\centering}|p{1.6cm}<{\centering}p{1.8cm}<{\centering}|p{1.6cm}<{\centering}}
				\hline
				& R3Live \cite{lin2022r} & Fast-LIVO \cite{zheng2022fast}     & Ours          \\ \hline
				$eee\_01$ & 1.69   & 0.28          & \textbf{0.21} \\
				$eee\_02$ & $\times$      & \textbf{0.18} & 0.23          \\
				$eee\_03$ & 0.64   & 0.26          & \textbf{0.22} \\
				$nya\_01$ & 0.63   & 0.34          & \textbf{0.18} \\
				$nya\_02$ & 0.35   & 0.29          & \textbf{0.19} \\
				$nya\_03$ & 0.23   & 0.29          & \textbf{0.20} \\
				$sbs\_01$ & 0.40   & 0.73          & \textbf{0.12} \\
				$sbs\_02$ & 0.27   & 0.25          & \textbf{0.22} \\
				$sbs\_03$ & \textbf{0.21}   & 0.24          & \textbf{0.21} \\ \hline
			\end{tabular}
	\end{center}
\end{table}

\begin{table}[]
	\begin{center}
	\caption{End to End Errors (unit: m)}
	\label{table2}
	\begin{tabular}{p{1.6cm}<{\centering}|p{1.6cm}<{\centering}p{1.8cm}<{\centering}|p{1.6cm}<{\centering}}
		\hline
		& R3Live \cite{lin2022r} & Fast-LIVO \cite{zheng2022fast}     & Ours           \\ \hline
		$r3live\_01$ & 0.097  & 0.070          & \textbf{0.021} \\
		$r3live\_03$ & 0.115  & 0.090          & \textbf{0.053} \\
		$r3live\_04$ & 0.071  & \textbf{0.050} & 0.120          \\
		$r3live\_05$ & 0.603  & \textbf{0.546} & \textbf{0.546} \\
		$r3live\_06$ & 0.036  & 13.463         & \textbf{0.024} \\ \hline
	\end{tabular}
	\end{center}
\end{table}

Results in Table \ref{table1} demonstrate that our system outperforms R3Live and Fast-LIVO for almost all sequences in terms of smaller ATE. "$\times$" means the system drifts halfway through the run, where our SR-LIVO has better robustness than R3Live on $NTU\_VIRAL$ dataset. The end-to-end errors are reported in Table \ref{table2}, whose overall trend is similar to the RMSE of ATE results. Our SR-LIVO achieves the lowest drift in four of the total five sequences. Fast-LIVO achieves large end-to-end error on sequence $hkust\_campus\_seq\_02$ because Fast-LIVO updates a single ESIKF with LiDAR and visual observations, while the low-quality visual observations would diverge the results of the filter.

\subsection{Comparison of LIO module and VIO module}
\label{Comparison of LIO module and VIO module}

Our system is designed based on the logic that: the state estimation performance of LIO module is more accurate than that of VIO module. Therefore, we provide quantitative data in this section to prove that our logical base is correct. We compare the ATE results of the VIO module with the LiDAR module on R3Live, Fast-LIVO and our SR-LIVO. Our proposed SR-LIVO does not include the VIO module, and we implement a VIO module modeled similar as R3Live to complete this ablation study.

\begin{table}[]
	\begin{center}
		\caption{RMSE of ATE Comparison on VIO module and LIO module (unit: m)}
		\label{table3}
		\begin{tabular}{c|p{0.7cm}<{\centering}p{0.7cm}<{\centering}|cc|cc}
			\hline
			& R3Live        & \begin{tabular}[c]{@{}c@{}}R3Live\\ (V)\end{tabular}     & \begin{tabular}[c]{@{}c@{}}Fast-\\ LIVO\end{tabular} & \begin{tabular}[c]{@{}c@{}}Fast-\\ LIVO(V)\end{tabular} & \begin{tabular}[c]{@{}c@{}}SR-\\ LIVO\end{tabular} & \begin{tabular}[c]{@{}c@{}}SR-\\ LIVO(V)\end{tabular}    \\ \hline
			$eee\_01$ & \textbf{1.69} & 1.71          & \textbf{0.28} & 0.30          & \textbf{0.21} & 0.24          \\
			$eee\_02$ & $\times$             & $\times$             & \textbf{0.18} & 0.26          & \textbf{0.23} & \textbf{0.23} \\
			$eee\_03$ & \textbf{0.64} & 0.81          & \textbf{0.26} & 0.27          & \textbf{0.22} & 0.27          \\
			$nya\_01$ & \textbf{0.63} & \textbf{0.63} & \textbf{0.33} & \textbf{0.33} & \textbf{0.18} & 0.19          \\
			$nya\_02$ & \textbf{0.35} & 0.41          & \textbf{0.29} & 0.30          & \textbf{0.19} & 0.20          \\
			$nya\_03$ & \textbf{0.23} & 0.32          & \textbf{0.22} & 0.29          & \textbf{0.20} & 0.24          \\
			$sbs\_01$ & \textbf{0.40} & 0.70          & 0.73          & \textbf{0.42} & \textbf{0.12} & 0.38          \\
			$sbs\_02$ & \textbf{0.27} & 0.33          & \textbf{0.25} & 0.31          & \textbf{0.22} & \textbf{0.22} \\
			$sbs\_03$ & \textbf{0.21} & 0.23          & \textbf{0.24} & 0.27          & \textbf{0.21} & 0.22          \\ \hline
		\end{tabular}
	\end{center}
\end{table}

The results in Table \ref{table3} demonstrate that: the accuracy of LIO modules is superior to that of VIO modules whether the framework is based on R3Live, Fast-LIVO or our SR-LIVO. This supports the conclusion that LIO is better at state estimation than VIO.

\subsection{Time Consumption}
\label{Time Consumption}

\begin{table}[]
	\begin{center}
		\caption{Time Consumption Per Sweep (unit: ms)}
		\label{table4}
		\begin{tabular}{c|cc|cc|cc}
			\hline
			& \multicolumn{2}{c|}{Vision}    & \multicolumn{2}{c|}{LiDAR} & \multicolumn{2}{c}{Total} \\ \hline
			& R3Live        & Ours           & R3Live   & Ours            & R3Live  & Ours            \\ \hline
			$r3live\_01$ & 38.90         & \textbf{20.86} & 17.51    & \textbf{9.67}   & 50.57   & \textbf{30.53}  \\
			$r3live\_02$ & 33.80         & \textbf{19.95} & 24.23    & \textbf{10.83}  & 49.95   & \textbf{30.78}  \\
			$r3live\_03$ & 37.16         & \textbf{21.51} & 18.16    & \textbf{9.55}   & 49.27   & \textbf{31.06}  \\
			$r3live\_04$ & 43.25         & \textbf{20.33} & 19.33    & \textbf{10.54}  & 56.14   & \textbf{30.87}  \\
			$r3live\_05$ & 40.23         & \textbf{19.64} & 20.95    & \textbf{11.78}  & 54.20   & \textbf{31.42}  \\
			$r3live\_06$ & 38.82         & \textbf{20.80} & 27.92    & \textbf{13.08}  & 57.43   & \textbf{33.88}  \\ \hline
			$eee\_01$    & \textbf{5.23} & 5.43           & 38.66    & \textbf{11.15}  & 31.02   & \textbf{16.58}  \\
			$eee\_02$    & $\times$             & \textbf{5.41}  & 30.04    & \textbf{9.69}   & 24.60   & \textbf{15.10}  \\
			$eee\_03$    & \textbf{5.00} & 6.93           & 27.43    & \textbf{10.00}  & 23.29   & \textbf{16.93}  \\
			$nya\_01$    & 6.58          & \textbf{6.02}  & 19.62    & \textbf{7.94}   & 19.67   & \textbf{13.96}  \\
			$nya\_02$    & \textbf{6.49} & 8.68           & 20.07    & \textbf{7.81}   & 19.87   & \textbf{16.49}  \\
			$nya\_03$    & \textbf{6.98} & 7.22           & 20.04    & \textbf{8.02}   & 20.34   & \textbf{15.24}  \\
			$sbs\_01$    & \textbf{5.10} & 5.69           & 24.30    & \textbf{8.97}   & 21.31   & \textbf{14.66}  \\
			$sbs\_02$    & 5.52          & \textbf{5.48}  & 26.74    & \textbf{9.31}   & 23.36   & \textbf{14.79}  \\
			$sbs\_03$    & 5.26          & \textbf{5.11}  & 26.47    & \textbf{10.24}  & 22.92   & \textbf{15.35}  \\ \hline
		\end{tabular}
	\end{center}
\end{table}

We evaluate the runtime breakdown (unit: ms) of our system and R3Live for all testing sequences. In general, the whole system framework consists of the VIO module and the LIO module. For each sequence, we test the time consumption of the above two modules, and the total time for handling a sweep. Results in Table \ref{table4} show that our system takes 30$\sim$34ms to handle a sweep on $R3Live$ dataset and 14$\sim$17ms to handle a sweep on $NTU\_VIRAL$ dataset, while R3Live takes 49$\sim$58ms to handle a sweep on $R3Live$ dataset and 10$\sim$31ms to handle a sweep on $NTU\_VIRAL$ dataset. That means our system can run around 1.6X faster than R3Live.

\subsection{Real-Time Performance Evaluation}
\label{Real-Time Performance Evaluation}

\begin{figure}
	\begin{center}
		\includegraphics[scale=0.5]{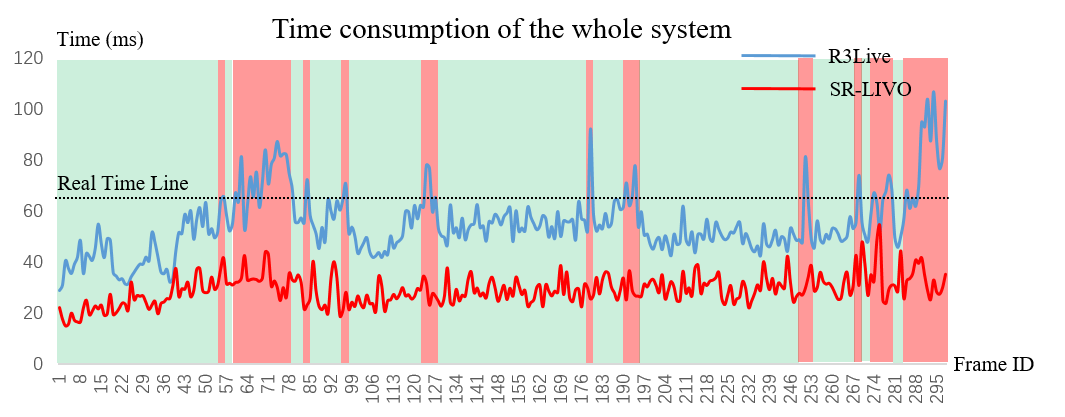}
		\caption{The curve of time consumption as the frame ID changes on sequence $r3live\_01$. R3Live cannot ensure the real-time performance for a considerable amount of time, while our SR-LIVO can run in real time stably.}
		\label{fig4}
	\end{center}
\end{figure}

We take $r3live\_01$ as the exemplar sequence, and plot the curve of time consumption as the frame ID changes. Fig. \ref{fig4} demonstrates that R3Live cannot ensure the real-time performance for a considerable amount of time, while our SR-LIVO can run in real time stably.

\subsection{Visualization for Map}
\label{Visualization for Map}

\begin{figure}
	\begin{center}
		\includegraphics[scale=0.62]{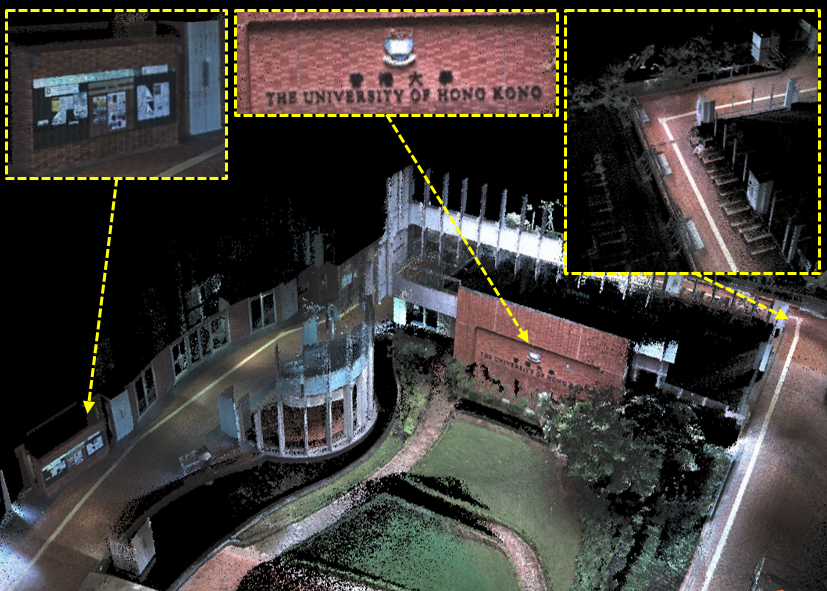}
		\caption{Our SR-LIVO is able to reconstruct a dense, 3D, RGB-colored point cloud map, which is comparable to the reconstruction result of R3Live (Fig. 1 in \cite{lin2022r}).}
		\label{fig5}
	\end{center}
\end{figure}

\begin{figure}
	\begin{center}
		\includegraphics[scale=0.42]{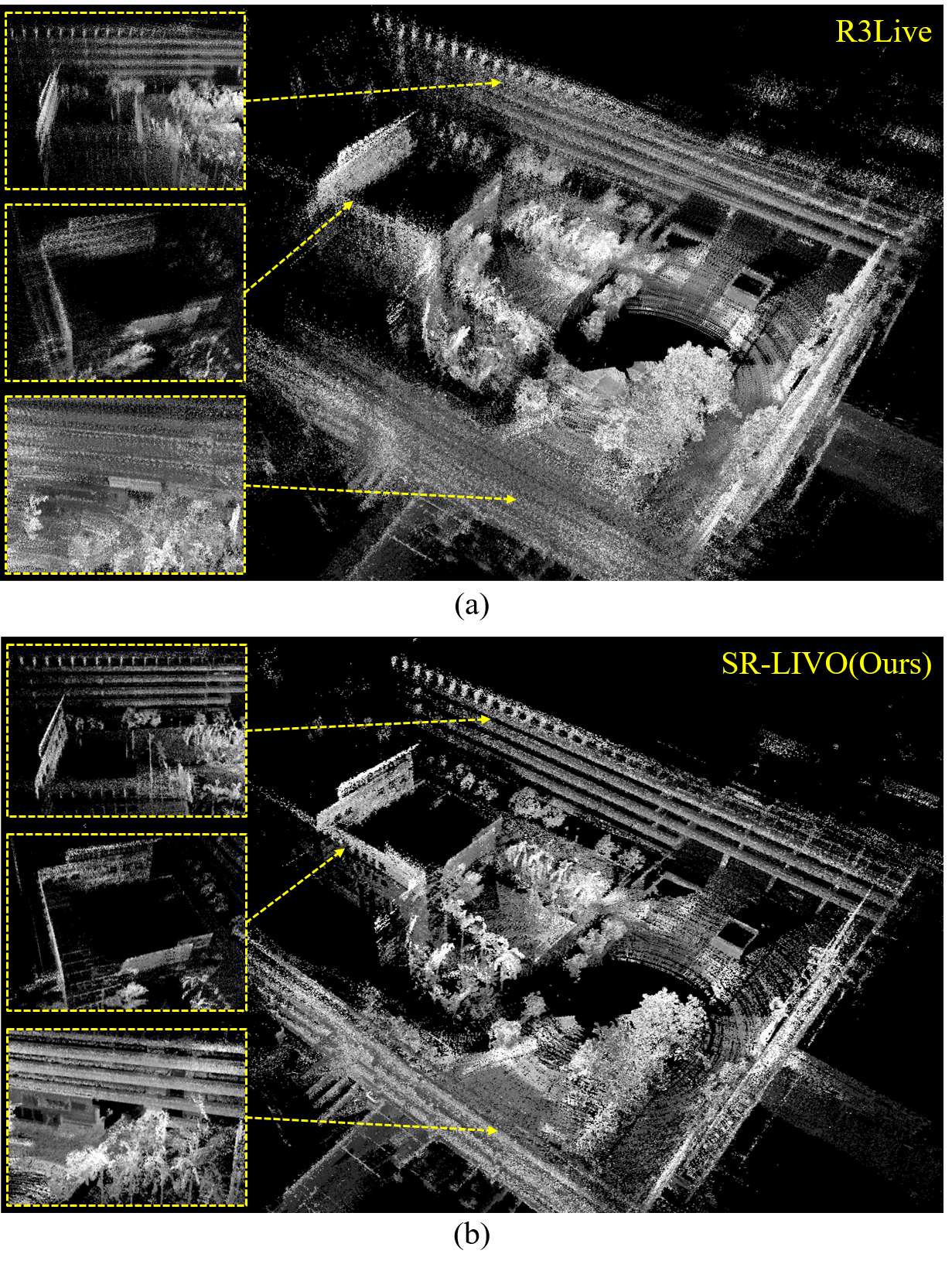}
		\caption{(a) and (b) are the reconstructed grayscale map on $eee\_01$ by R3Live and SR-LIVO respectively, while our system achieves significantly better reconstruction result.}
		\label{fig6}
	\end{center}
\end{figure}

Fig. \ref{fig5} shows the ability of our SR-LIVO to reconstruct a dense, 3D, RGB-colored point cloud map on the exemplar sequence (e.g., $r3live\_01$), which is comparable to the reconstruction result of R3Live (Fig. 1 in \cite{lin2022r}). Under the premise that the reconstruction result is equal, our method can run stably in real time while R3Live cannot, demonstrating the strength of our approach.

The camera images of $NTU\_VIRAL$ dataset are grayscale images, leading to the grayscale map. Fig. \ref{fig6} compares the visualizations of our reconstructed grayscale map with R3Live on the exemplar sequence (e.g., $eee\_01$) of $NTU\_VIRAL$ dataset, on which our system achieves significantly better reconstruction result.

\section{Conclusion}
\label{Conclusion}

This paper proposes a novel LIV-SLAM system, named SR-LIVO, which adapts the sweep reconstruction method to align the end timestamp of reconstructed sweep to the timestamp of captured image. Thus, the state of all image-captured moments can be solved by the more reliable LIO module instead of the hypersensitive VIO module. In SR-LIVO, we utilize an ESIKF to solve state in LIO module, and utilize an ESIKF to optimize camera parameters in vision module respectively for optimal state estimation and colored point cloud map reconstruction.

Our system achieves state-of-the-art pose accuracy on two public datasets, and achieves much lower time consumption than R3Live while keeping the reconstruction result comparable or better. Future work includes adding loop closing module to this framework.

\bibliographystyle{IEEEtrans}
\bibliography{IEEEabrv,IEEEExample}

\begin{thebibliography}{10}
\providecommand{\url}[1]{#1}
\csname url@samestyle\endcsname
\providecommand{\newblock}{\relax}
\providecommand{\bibinfo}[2]{#2}
\providecommand{\BIBentrySTDinterwordspacing}{\spaceskip=0pt\relax}
\providecommand{\BIBentryALTinterwordstretchfactor}{4}
\providecommand{\BIBentryALTinterwordspacing}{\spaceskip=\fontdimen2\font plus
\BIBentryALTinterwordstretchfactor\fontdimen3\font minus
  \fontdimen4\font\relax}
\providecommand{\BIBforeignlanguage}[2]{{%
\expandafter\ifx\csname l@#1\endcsname\relax
\typeout{** WARNING: IEEEtranS.bst: No hyphenation pattern has been}%
\typeout{** loaded for the language `#1'. Using the pattern for}%
\typeout{** the default language instead.}%
\else
\language=\csname l@#1\endcsname
\fi
#2}}
\providecommand{\BIBdecl}{\relax}
\BIBdecl

\bibitem{dellenbach2022ct}
P.~Dellenbach, J.-E. Deschaud, B.~Jacquet, and F.~Goulette, ``Ct-icp: Real-time
  elastic lidar odometry with loop closure,'' in \emph{2022 International
  Conference on Robotics and Automation (ICRA)}.\hskip 1em plus 0.5em minus
  0.4em\relax IEEE, 2022, pp. 5580--5586.

\bibitem{gao2019flying}
F.~Gao, W.~Wu, W.~Gao, and S.~Shen, ``Flying on point clouds: Online trajectory
  generation and autonomous navigation for quadrotors in cluttered
  environments,'' \emph{Journal of Field Robotics}, vol.~36, no.~4, pp.
  710--733, 2019.

\bibitem{kong2021avoiding}
F.~Kong, W.~Xu, Y.~Cai, and F.~Zhang, ``Avoiding dynamic small obstacles with
  onboard sensing and computation on aerial robots,'' \emph{IEEE Robotics and
  Automation Letters}, vol.~6, no.~4, pp. 7869--7876, 2021.

\bibitem{levinson2011towards}
J.~Levinson, J.~Askeland, J.~Becker, J.~Dolson, D.~Held, S.~Kammel, J.~Z.
  Kolter, D.~Langer, O.~Pink, V.~Pratt \emph{et~al.}, ``Towards fully
  autonomous driving: Systems and algorithms,'' in \emph{2011 IEEE intelligent
  vehicles symposium (IV)}.\hskip 1em plus 0.5em minus 0.4em\relax IEEE, 2011,
  pp. 163--168.

\bibitem{lin2022r}
J.~Lin and F.~Zhang, ``R 3 live: A robust, real-time, rgb-colored,
  lidar-inertial-visual tightly-coupled state estimation and mapping package,''
  in \emph{2022 International Conference on Robotics and Automation
  (ICRA)}.\hskip 1em plus 0.5em minus 0.4em\relax IEEE, 2022, pp.
  10\,672--10\,678.

\bibitem{lin2021r}
J.~Lin, C.~Zheng, W.~Xu, and F.~Zhang, ``R 2 live: A robust, real-time,
  lidar-inertial-visual tightly-coupled state estimator and mapping,''
  \emph{IEEE Robotics and Automation Letters}, vol.~6, no.~4, pp. 7469--7476,
  2021.

\bibitem{nguyen2022ntu}
T.-M. Nguyen, S.~Yuan, M.~Cao, Y.~Lyu, T.~H. Nguyen, and L.~Xie, ``Ntu viral: A
  visual-inertial-ranging-lidar dataset, from an aerial vehicle viewpoint,''
  \emph{The International Journal of Robotics Research}, vol.~41, no.~3, pp.
  270--280, 2022.

\bibitem{peng2023vilo}
G.~Peng, Y.~Zhou, L.~Hu, L.~Xiao, Z.~Sun, Z.~Wu, and X.~Zhu, ``Vilo slam:
  Tightly coupled binocular vision--inertia slam combined with lidar,''
  \emph{Sensors}, vol.~23, no.~10, p. 4588, 2023.

\bibitem{qinonline}
T.~Qin and S.~Shen, ``Online temporal calibration for monocular visual-inertial
  systems. in 2018 ieee,'' in \emph{RSJ International Conference on Intelligent
  Robots and Systems (IROS)}, pp. 3662--3669.

\bibitem{shan2021lvi}
T.~Shan, B.~Englot, C.~Ratti, and D.~Rus, ``Lvi-sam: Tightly-coupled
  lidar-visual-inertial odometry via smoothing and mapping,'' in \emph{2021
  IEEE international conference on robotics and automation (ICRA)}.\hskip 1em
  plus 0.5em minus 0.4em\relax IEEE, 2021, pp. 5692--5698.

\bibitem{shao2019stereo}
W.~Shao, S.~Vijayarangan, C.~Li, and G.~Kantor, ``Stereo visual inertial lidar
  simultaneous localization and mapping,'' in \emph{2019 IEEE/RSJ International
  Conference on Intelligent Robots and Systems (IROS)}.\hskip 1em plus 0.5em
  minus 0.4em\relax IEEE, 2019, pp. 370--377.

\bibitem{wang2021dv}
W.~Wang, J.~Liu, C.~Wang, B.~Luo, and C.~Zhang, ``Dv-loam: Direct visual lidar
  odometry and mapping,'' \emph{Remote Sensing}, vol.~13, no.~16, p. 3340,
  2021.

\bibitem{wang2022mvil}
Y.~Wang and H.~Ma, ``mvil-fusion: Monocular visual-inertial-lidar simultaneous
  localization and mapping in challenging environments,'' \emph{IEEE Robotics
  and Automation Letters}, vol.~8, no.~2, pp. 504--511, 2022.

\bibitem{yin2023solid}
T.~Yin, J.~Yao, Y.~Lu, and C.~Na, ``Solid-state-lidar-inertial-visual odometry
  and mapping via quadratic motion model and reflectivity information,''
  \emph{Electronics}, vol.~12, no.~17, p. 3633, 2023.

\bibitem{yuan2022sr}
Z.~Yuan, F.~Lang, T.~Xu, and X.~Yang, ``Sr-lio: Lidar-inertial odometry with
  sweep reconstruction,'' \emph{arXiv preprint arXiv:2210.10424}, 2022.

\bibitem{yuan2023sdv}
Z.~Yuan, Q.~Wang, K.~Cheng, T.~Hao, and X.~Yang, ``Sdv-loam: Semi-direct
  visual-lidar odometry and mapping,'' \emph{IEEE Transactions on Pattern
  Analysis and Machine Intelligence}, 2023.

\bibitem{zhang2018laser}
J.~Zhang and S.~Singh, ``Laser--visual--inertial odometry and mapping with high
  robustness and low drift,'' \emph{Journal of field robotics}, vol.~35, no.~8,
  pp. 1242--1264, 2018.

\bibitem{zheng2022fast}
C.~Zheng, Q.~Zhu, W.~Xu, X.~Liu, Q.~Guo, and F.~Zhang, ``Fast-livo: Fast and
  tightly-coupled sparse-direct lidar-inertial-visual odometry,'' in \emph{2022
  IEEE/RSJ International Conference on Intelligent Robots and Systems
  (IROS)}.\hskip 1em plus 0.5em minus 0.4em\relax IEEE, 2022, pp. 4003--4009.

\bibitem{zuo2019lic}
X.~Zuo, P.~Geneva, W.~Lee, Y.~Liu, and G.~Huang, ``Lic-fusion:
  Lidar-inertial-camera odometry,'' in \emph{2019 IEEE/RSJ International
  Conference on Intelligent Robots and Systems (IROS)}.\hskip 1em plus 0.5em
  minus 0.4em\relax IEEE, 2019, pp. 5848--5854.

\bibitem{zuo2020lic}
X.~Zuo, Y.~Yang, P.~Geneva, J.~Lv, Y.~Liu, G.~Huang, and M.~Pollefeys,
  ``Lic-fusion 2.0: Lidar-inertial-camera odometry with sliding-window
  plane-feature tracking,'' in \emph{2020 IEEE/RSJ International Conference on
  Intelligent Robots and Systems (IROS)}.\hskip 1em plus 0.5em minus
  0.4em\relax IEEE, 2020, pp. 5112--5119.

\end{thebibliography}




\ifCLASSOPTIONcaptionsoff
  \newpage
\fi



%

%


\begin{IEEEbiography}[{\includegraphics[width=1in,height=1.25in,clip,keepaspectratio]{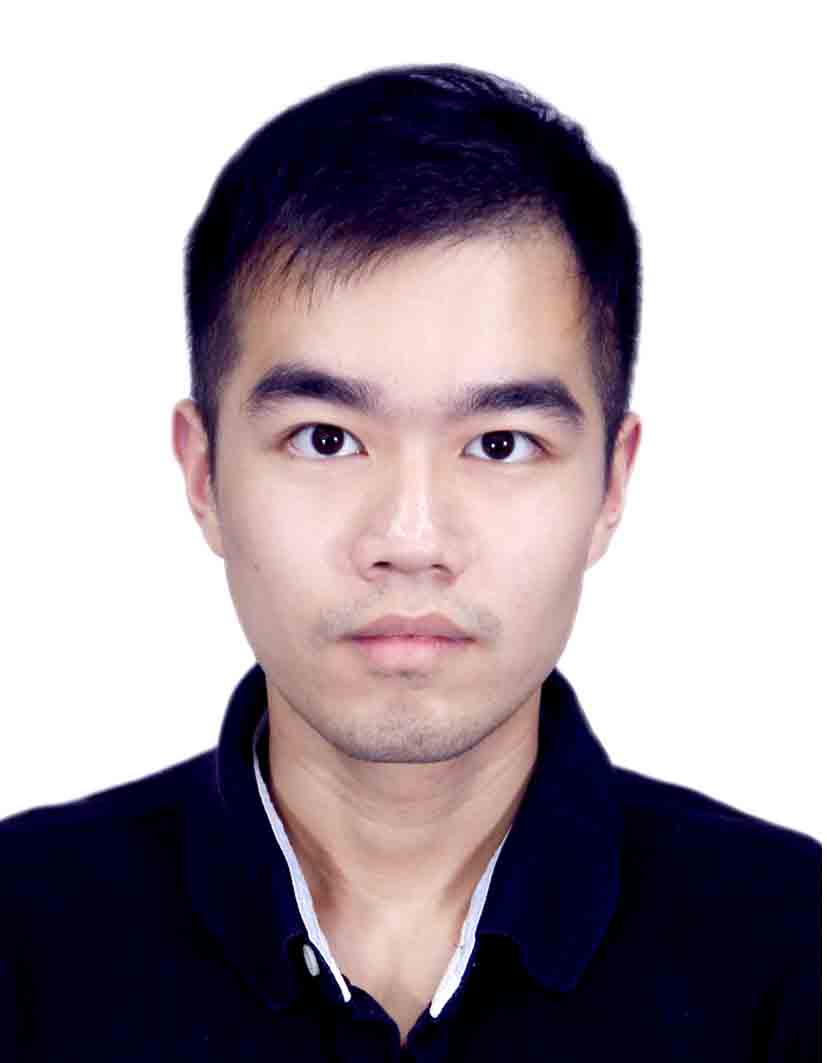}}]{Zikang~Yuan}
	received the B.E. degree from Huazhong University of Science and Technology (HUST), Wuhan, China, in 2018. He is currently a 4th year PhD student of HUST, School of Institute of Artificial Intelligence. He has published two papers on ACM MM, three papers on TMM, one paper on TPAMI and one paper on IROS. His research interests include monocular dense mapping, RGB-D simultaneous localization and mapping, visual-inertial state estimation, LiDAR-inertial state estimation, LiDAR-inertial-wheel state estimation and visual-LiDAR pose estimation and mapping.
\end{IEEEbiography}

\vspace{-1.5cm}

\begin{IEEEbiography}[{\includegraphics[width=1in,height=1.25in,clip,keepaspectratio]{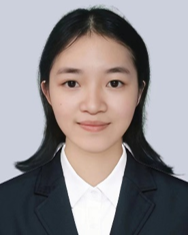}}]{Jie~Deng}
	is currently a 4th year B.E. student of HUST, School of Electronic Information and Communications. Her research interests include LiDAR-inertial state estimation, LiDAR-inertial-wheel state estimation and visual-LiDAR pose estimation and mapping.
\end{IEEEbiography}

\vspace{-1.5cm}

\begin{IEEEbiography}[{\includegraphics[width=1in,height=1.25in,clip,keepaspectratio]{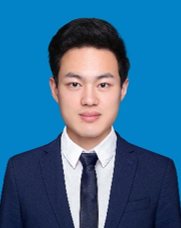}}]{Ruiye~Ming}
	received the B.E. degree from Zhengzhou University(ZZU), Zhengzhou, China, in 2022. He is currently a 2nd year graduate student of HUST, School of Electronic Information and  Communications. His research interests include visual-LiDAR odometry, LiDAR point based loop closing, object reconstruction and deep-learning based depth estimation.
\end{IEEEbiography}

\vspace{-1.5cm}

\begin{IEEEbiography}[{\includegraphics[width=1in,height=1.25in,clip,keepaspectratio]{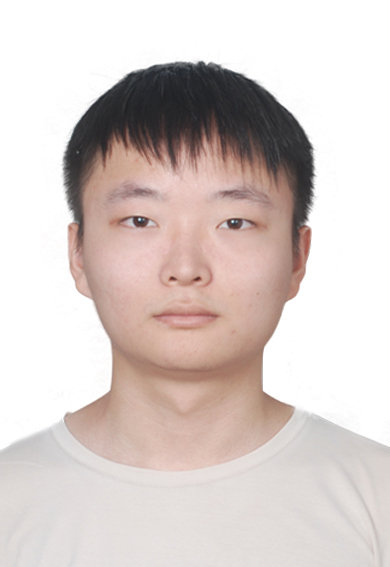}}]{Fengtian~Lang}
	received the B.E. degree from Huazhong University of Science and Technology (HUST), Wuhan, China, in 2023. He is currently a 1st year graduate student of HUST, School of Electronic Information and  Communications. He has published one paper on IROS. His research interests include LiDAR-inertial state estimation, LiDAR-inertial-wheel state estimation and LiDAR point based loop closing.
\end{IEEEbiography}

\vspace{-1.5cm}

\begin{IEEEbiography}[{\includegraphics[width=1in,height=1.25in,clip,keepaspectratio]{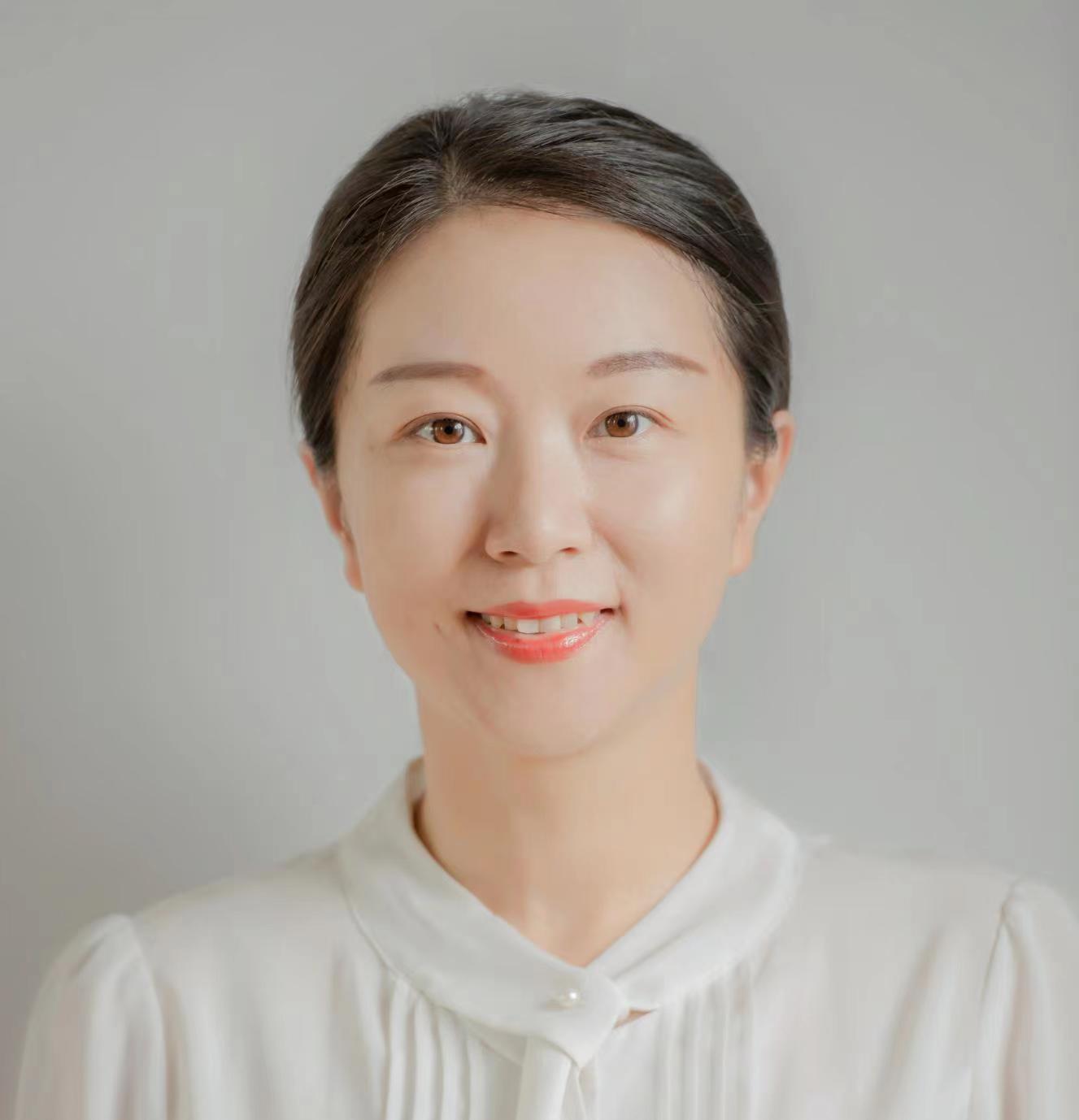}}]{Xin~Yang}
	received her PhD degree in University
	of California, Santa Barbara in 2013. She worked
	as a Post-doc in Learning-based Multimedia Lab at
	UCSB (2013-2014). She is current Professor
	of Huazhong University of Science and Technology
	School of Electronic Information and Communications.
	Her research interests include
	simultaneous localization and mapping, augmented
	reality, and medical image analysis. She has published
	over 90 technical papers, including TPAMI, IJCV,
	TMI, MedIA, CVPR, ECCV, MM, etc., co-authored two books and holds 3 U.S. Patents. Prof.
	Yang is a member of IEEE and a member of ACM.
\end{IEEEbiography}




\end{document}